\documentclass{article}
\usepackage{spconf,amsmath,graphicx}
\usepackage{booktabs}
\usepackage{multirow}
\usepackage{stfloats}
\usepackage{xcolor,xspace}
\usepackage{bm,bbm,amsfonts}
\usepackage{etoolbox}
\usepackage{hyperref}

\AtBeginDocument{
  
  \newcommand{\argmin}{\operatornamewithlimits{arg\,min}}
  \newcommand{\argmax}{\operatornamewithlimits{arg\,max}}
}
\DeclareMathOperator*{\diag}{diag}
\newcommand{\heading}[2][0.8ex]{\vspace{#1}\noindent\textbf{#2}\hspace{0.2em}}

\patchcmd{\thebibliography}
  {\list}
  {\small\list}
  {}
  {}

\makeatletter
\DeclareRobustCommand\onedot{\futurelet\@let@token\@onedot}
\def\@onedot{\ifx\@let@token.\else.\null\fi\xspace}

\def\ie{\emph{i.e}\onedot}

\def\etal{\emph{et al}\onedot}
\makeatother

\title{Uncertainty-Guided Latent Diffusion Models \\ for Faithful Super Resolution}
\twoauthors
 {Ren Wang}
	{National Taiwan University}
 {Yung-Yu Chuang
 \thanks{\hspace*{-1.8em}{\footnotesize{Thanks to NSTC, NTU, and MoE for their support through grants 113-2221-E-002-112-MY3, 114L892203 and Taiwan Centers of Excellence (TCE).}}}
 }
	{National Taiwan University, NTU AI-CoRE}

\begin{document}
\maketitle

\begin{abstract}
The perception-distortion trade-off poses a fundamental challenge in single-image super-resolution (SR). Although diffusion-based SR methods excel at generating perceptually realistic images, achieving high fidelity remains a key limitation. Recent advances in diffusion-based SR have shown promise in improving fidelity, but these methods often compromise perceptual quality due to their high reliance on a high-fidelity image. To address this, we introduce \emph{UGDiff}, a novel diffusion guidance paradigm designed to further improve the perception-distortion balance. In particular, we first estimate the reconstruction uncertainty of the latent features corresponding to a high-fidelity image. This uncertainty is then used to guide the diffusion process to selectively restore high-frequency details in high-uncertainty regions, while preserving fidelity elsewhere. Furthermore, our guidance method adaptively identifies the high-uncertainty regions by considering not only the estimated uncertainty but also the posterior variance of the diffusion sampler at each timestep. This relaxes the reliance on the high-fidelity image in the later stages of sampling, thereby achieving a better perception-distortion balance. Extensive experimental results demonstrate that our method performs favorably against state-of-the-art diffusion-based SR methods.
\end{abstract}

\section{Introduction}
\label{sec:intro}

Single-image super-resolution (SR) aims to recover a high-resolution image $\mathbf{I}_\text{hq}$ from a low-resolution observation $\mathbf{I}_\text{lq}$. Typically, the problem can be formulated as
\begin{align}
    \mathbf{I}_\text{hq} = \argmin\limits_{\mathbf{x}} \frac{1}{2\sigma^2}\left\| \mathbf{I}_\text{lq} - \mathcal{H}(\mathbf{x}) \right\|^2_2 + \lambda \mathcal{P}(\mathbf{x}),
\end{align}
where $\mathcal{H}$ is the degradation model, $\sigma$ is the associated noise level, and $\mathcal{P}$ is the image prior. The main challenge in optimizing this objective is that there are many possible solutions; some tend towards high-fidelity (\ie, optimizing the first term), while others lean towards perceptual quality (\ie, optimizing the second term). This is the well-known perception-distortion trade-off~\cite{pd_tradeoff}. The motivation of this work is to achieve a better perceptual-distortion balance rather than merely optimizing for a single metric. Specifically, we refer to this objective as \emph{faithful super-resolution}.

Recently, diffusion model-based SR methods~\cite{sr3, srdiff, resshift, stablesr} have shown great capability of generating realistic high-frequency details to improve perceptual quality. However, achieving high fidelity is challenging because of the stochastic nature of diffusion models. Several prior studies have tried to address this problem. PASD~\cite{pasd} uses ControlNet~\cite{controlnet} to better learn the guidance from the input image. DiffBIR~\cite{diffbir} proposes steering the diffusion process towards a high-fidelity image through image gradient weighting. SUPIR~\cite{supir} proposes a restoration-guided sampling that performs a weighted interpolation between the sampling point and the input image at each step. DADiff~\cite{dadiff} integrates a regression-based restoration model into diffusion guidance. PiSA-SR~\cite{pisa-sr} blends the outputs of pixel-level and semantic-level enhancement modules in the latent space. FaithDiff~\cite{faithdiff} proposes a feature alignment module to identify useful information from the input image. However, these methods often trade perceptual quality for fidelity due to their high reliance on the input image or a high-fidelity image.

\begin{figure}[t]
  \centering
   \includegraphics[width=0.75\linewidth]{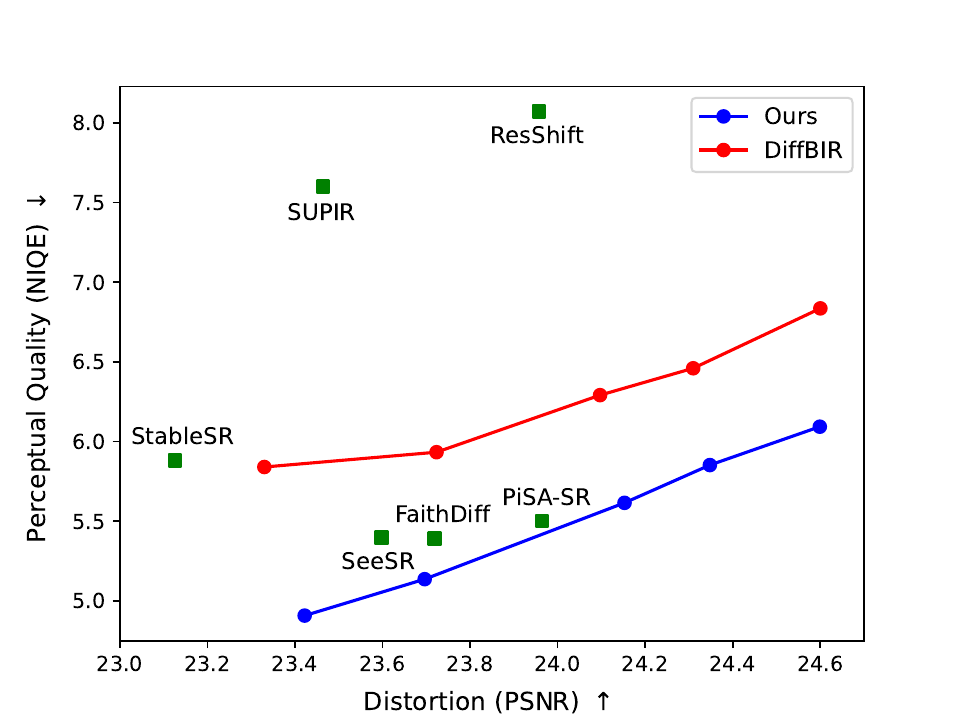}
   \caption{Perception-Distortion trade-off of diffusion-based SR methods on RealSR. The DiffBIR curve corresponds to various values of $s$, while ours varies $\gamma$ at a fixed $s=100$.}
   \label{fig:tradeoff}
\end{figure}

In this work, we leverage uncertainty estimation for faithful super-resolution in latent diffusion models (LDMs)~\cite{sd}. While the VAE~\cite{vae} in LDMs also estimates uncertainty, it is typically ignored in practice due to the negligible KL-divergence loss during training. Consequently, we introduce a specialized encoder to estimate the uncertainty of a high-fidelity image in the latent space. Following Han~\etal~\cite{han2025uncertainty}, we propose a simple but effective $\mathcal{L}_2$ loss to train the encoder. This uncertainty is then used to guide the diffusion process to selectively restore high-frequency details in high-uncertainty regions, while preserving fidelity elsewhere. Furthermore, we adaptively identify the high-uncertainty regions by incorporating not only the estimated uncertainty but also the posterior variance of the diffusion sampler. This enables our method to relax the reliance on the high-fidelity image in the later stages of sampling, thus improving the perception-distortion balance. Fig.~\ref{fig:tradeoff} shows that our method achieves the best balance among all of the compared diffusion-based SR methods.

Our contributions are summarized as follows:

\begin{itemize}
    \item We propose a simple yet effective method to train an uncertainty estimation model in the latent space.
    \item We propose a novel guidance method to make the diffusion model selectively and adaptively restore high-frequency details in high-uncertainty regions.
    \item Extensive experiments demonstrate that our method strikes a better perception-distortion balance compared to state-of-the-art diffusion-based SR methods.
\end{itemize}

\begin{figure*}
    \centering
    \includegraphics[width=0.85\linewidth]{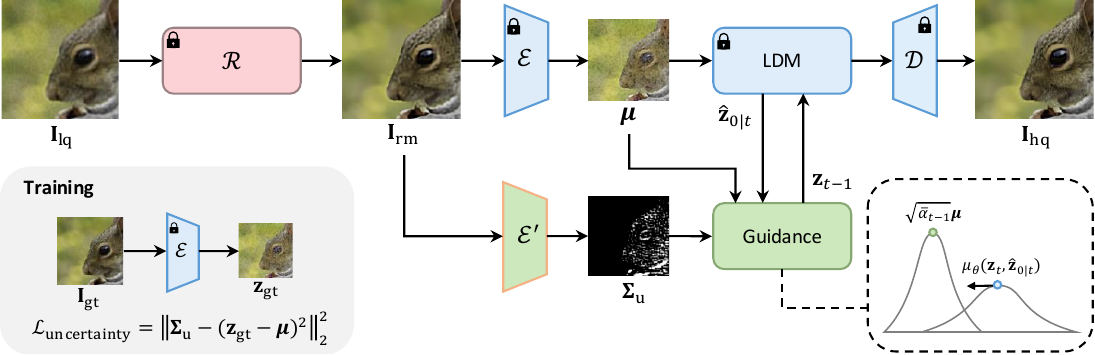}
    \caption{Overview of our method. Given a high-fidelity image $\mathbf{I}_\text{rm}$ restored from the low-quality image $\mathbf{I}_\text{lq}$ by a regression-based restoration model $\mathcal{R}$, our uncertainty estimation model $\mathcal{E}'$ produces a variance map $\mathbf{\Sigma}_\text{u}$ specifically for $\bm{\mu}=\mathcal{E}(\mathbf{I}_\text{rm})$. During the reverse process of a pretrained latent diffusion model, we rectify $\mu_\theta(\mathbf{z}_t, \hat{\mathbf{z}}_{0|t})$, \ie, the mean of the posterior distribution $p(\mathbf{z}_{t-1}|\mathbf{z}_t, \hat{\mathbf{z}}_{0|t})$, at each timestep $t$. Our guidance aims to push $\mu_\theta(\mathbf{z}_t, \hat{\mathbf{z}}_{0|t})$ to the direction of $\bm{\mu}$ according to the uncertainty at each pixel. Repeating this process, we get the final restored image $\mathbf{I}_\text{hq}$ at timestep $0$, which is expected to have plentiful details in textured regions while remaining robust against hallucinations and artifacts elsewhere.}
    \label{fig:overview}
\end{figure*}

\section{Related work}
\label{sec:related}

\subsection{Diffusion-Based Image Super Resolution}

SR3~\cite{sr3} proposes to train a conditional diffusion model from scratch for super-resolution. SRDiff~\cite{srdiff} and ResShift~\cite{resshift} propose to train diffusion models in residual space. IR-SDE~\cite{irsde} adds low-quality images to the initial noise during training, and lets the model be directly conditioned on the noised data at each step. StableSR~\cite{stablesr} utilizes Stable Diffusion~\cite{sd} as a strong prior for generating high-frequency information. PASD~\cite{pasd} leverages ControlNet~\cite{controlnet} to better learn the guidance from input images. DiffBIR~\cite{diffbir} uses a restored image instead of the input image as the condition of ControlNet, and proposes a restoration guidance to improve fidelity. SeeSR~\cite{seesr} and SUPIR~\cite{supir} show that text prompts offer effective semantic guidance for super-resolution. FaithDiff~\cite{faithdiff} proposes replacing ControlNet with a feature alignment module.
OSEDiff~\cite{osediff} and PiSA-SR~\cite{pisa-sr} apply LoRA~\cite{lora} to achieve one-step sampling for super-resolution.
In this work, we choose DiffBIR as the baseline to build our method.

\subsection{Uncertainty in Image Super Resolution}

Kendall and Gal~\cite{bayes_uncertainty} give a fundamental framework of uncertainty in deep learning, where classification problems are based on entropy and regression problems are based on Gaussian likelihood. UDL~\cite{udl} and UGAN~\cite{ugan} use uncertainty to identify the challenging pixels of an image and prioritize them for optimization. Liu~\etal~\cite{spectral_uncertainty} use a similar framework, but operate in the frequency domain. Fang~\etal~\cite{uncertainty_kernel} incorporate uncertainty learning into kernel estimation to improve the robustness of MAP-based SR. Han~\etal~\cite{han2025uncertainty} propose to train an uncertainty estimation model using optimal uncertainty when the mean is known, and integrate it into a VAE~\cite{vae}-like framework for stochastic SR. UPSR~\cite{upsr} proposes uncertainty-guided noise weighting to enable region-specific noise level control, given that slight noise is more advantageous for reconstructing flat areas, and vice versa. BUFF~\cite{buff} employs a Bayesian network to generate an uncertainty mask for diffusion models, which is used to weight the noise during training and condition the denoiser. Our work draws significant inspiration from Han~\etal~\cite{han2025uncertainty}; however, they neither used a diffusion model nor performed operations in the latent space.

\section{Method}

\subsection{Preliminaries}
\label{sec:preliminaries}

We follow the diffusion formulation adopted by LDM~\cite{sd}. Given a real image $\mathbf{x}_0 \sim q(\mathbf{x})$ and a pretrained encoder $\mathcal{E}$, LDM applies the forward process of DDPM~\cite{DDPM} to the clean latent code $\mathbf{z}_0 = \mathcal{E}(\mathbf{x}_0)$ for $t \in \{1, 2, ..., T\}$ by
\begin{align}
    q(\mathbf{z}_{t}|\mathbf{z}_{0}) = \mathcal N(\mathbf{z}_{t};\sqrt{\bar{\alpha}_{t}}\mathbf{z}_{0}, (1-\bar{\alpha}_{t})\mathbf{I}),
\label{eq:forward}
\end{align}
where $\mathbf{z}_t$ is the noised latent code, $\bar{\alpha}_{t} = \prod_{i=1}^{t}(1-\beta_{i})$, and $\beta_t$ is the pre-defined variance at timestep $t$. Then, starting from $\mathbf{z}_T \sim \mathcal{N}(\mathbf{0}, \mathbf{I})$, the reverse process is given by
\begin{align}
    p(\mathbf{z}_{t-1}|\mathbf{z}_t, \hat{\mathbf{z}}_{0|t}) = \mathcal N(\mathbf{z}_{t-1};\mu_\theta(\mathbf{z}_{t}, \hat{\mathbf{z}}_{0|t}), \sigma_t^2 \mathbf{I}),
\label{eq:reverse}
\end{align}
where
\begin{align}
    \mu_\theta(\mathbf{z}_t, \hat{\mathbf{z}}_{0|t}) &= \frac{\sqrt{\alpha_t}(1-\bar{\alpha}_{t-1})}{1-\bar{\alpha}_t}\mathbf{z}_t + \frac{\sqrt{\bar{\alpha}_{t-1}}\beta_t}{1-\bar{\alpha}_t} \hat{\mathbf{z}}_{0|t}, \nonumber \\
    \sigma_t^2 &= \frac{1-\bar{\alpha}_{t-1}}{1-\bar{\alpha}_t} \beta_t,
\end{align}
and
\begin{align}
    \hat{\mathbf{z}}_{0|t} = \frac{1}{\sqrt{\bar{\alpha}_t}} \left( \mathbf{x}_t - \sqrt{1-\bar{\alpha}_t} \epsilon_\theta(\mathbf{z}_t, t) \right),
\label{eq:z_0t}
\end{align}
where $\epsilon_\theta(\mathbf{z}_t, t)$ is the noise estimated by an LDM model $\theta$. Finally, we can obtain $\hat{\mathbf{x}}_0=\mathcal{D}(\hat{\mathbf{z}}_{0|1}) \sim q(\mathbf{x})$ given a pretrained decoder $\mathcal{D}$. For SR, $\mu_\theta$ is typically conditioned on $\mathcal{E}(\mathbf{I}_\text{lq})$, \ie, the estimated noise is $\epsilon_\theta(\mathbf{z}_t, \mathcal{E}(\mathbf{I}_\text{lq}), t)$.

\subsection{Uncertainty Estimation in Latent Space}
\label{sec:uncertainty}

Recall that while the pretrained encoder $\mathcal{E}$ also estimates uncertainty, it is typically ignored in practice due to the negligible KL-divergence loss during training. In this section, we describe how we train our uncertainty estimation model $\mathcal{E}'$ in the latent space. Fig.~\ref{fig:overview} gives an overview of our method. For simplicity, we let $\bm{\mu}, \mathbf{\Sigma}_\text{u} \in \mathbb{R}^d$ respectively denote the mean and the diagonal entries of the covariance matrix, where $d=H\times W \times C$. As in ~\cite{bayes_uncertainty}, we can train $\mathcal{E}'$ by minimizing the Gaussian negative log-likelihood (NLL) as
\begin{align}
    \mathcal{L}_\text{nll} = \mathbb{E} \left[ \sum\limits_{i=1}^d \frac{( \mathbf{z}_{\text{gt}, i} - {\bm{\mu}}_i )^2}{2\mathbf{\Sigma}_{\text{u},i}} + \frac{1}{2}\ln\mathbf{\Sigma}_{\text{u},i} \right],
\label{eq:loss_nll}
\end{align}
where $\mathbf{z}_\text{gt}=\mathcal{E}(\mathbf{I}_\text{gt})$ for the ground-truth image $\mathbf{I}_\text{gt}$. Typically, both $\bm{\mu}$ and $\mathbf{\Sigma}_\text{u}$ are estimated by the same model, requiring a non-trivial balance between the two loss terms during training. Inspired by Han~\etal~\cite{han2025uncertainty}, if $\bm{\mu}$ is estimated by an existing regression-based restoration model $\mathcal{R}$, \ie, $\bm{\mu}=\mathcal{E}(\mathcal{R}(\mathbf{I}_\text{lq}))=\mathcal{E}(\mathbf{I}_\text{rm})$, we can have the optimal uncertainty as
\begin{align}
    \mathbf{\Sigma}^\star_\text{u} &= \argmin\limits_{\mathbf{\Sigma}_\text{u}} \mathcal{L}_\text{nll} = (\mathbf{z}_\text{gt} - \bm{\mu}) \odot (\mathbf{z}_\text{gt} - \bm{\mu}).
\label{eq:var_star}
\end{align}
Then, we can train $\mathcal{E}'$ by a simple $\mathcal{L}_2$ loss as
\begin{align}
    \mathcal{L}_\text{uncertainty} = \mathbb{E} \left[ \left\| \mathbf{\Sigma}_\text{u} - \mathbf{\Sigma}^\star_\text{u} \right\|^2_2 \right],
\label{eq:loss_uncertainty}
\end{align}
where $\mathbf{\Sigma}_\text{u}=\mathcal{E}'(\mathbf{I}_\text{rm})$. With this simple training objective, our model can focus on uncertainty estimation without compromising mean accuracy.

\begin{figure*}[t]
  \centering
   \includegraphics[width=0.80\linewidth]{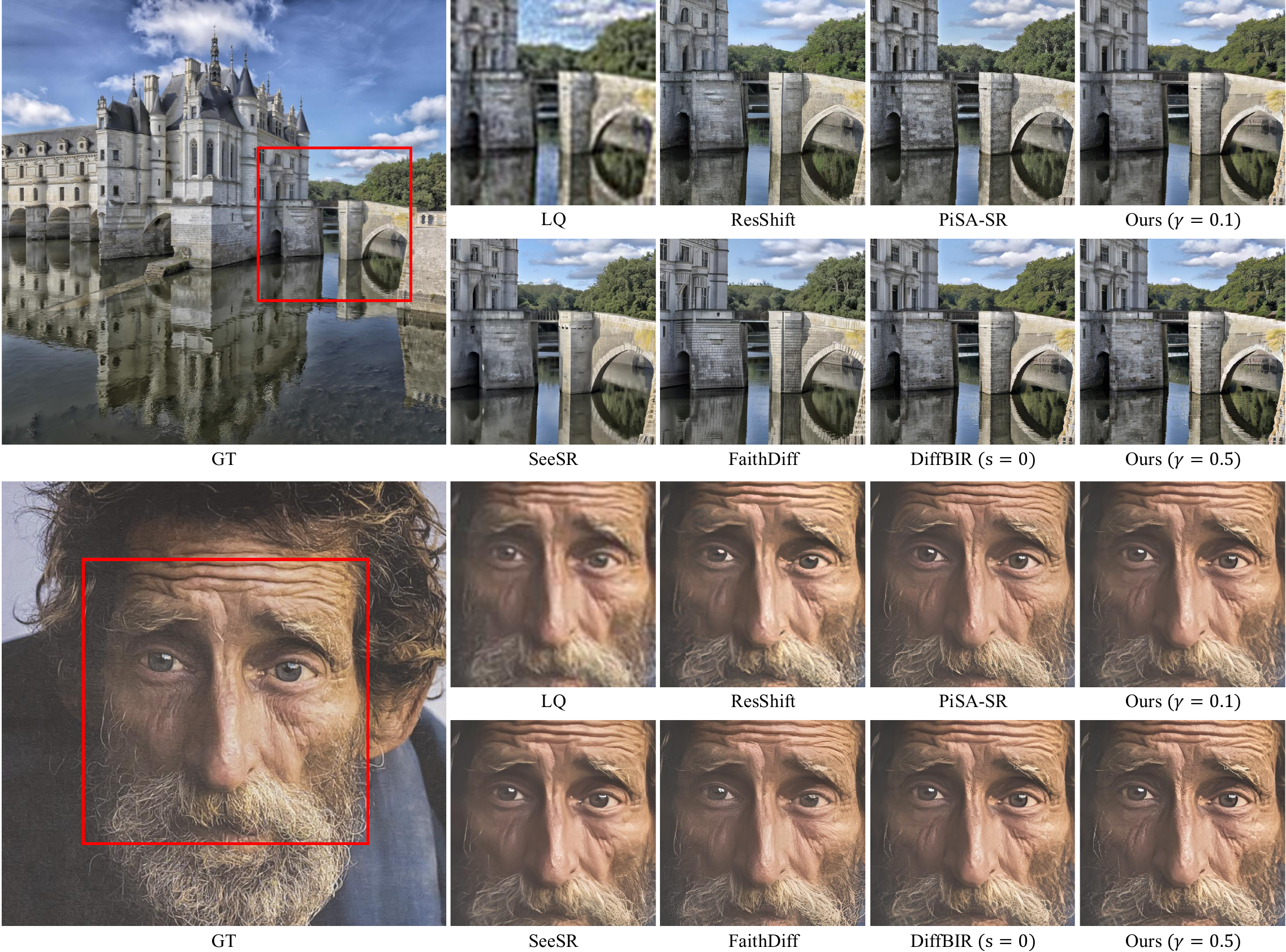}
   \caption{Comparison of $4 \times$ upsampling on DIV2K-Val (upper) and RealSR (bottom). We set $s=100$ for all of our methods.}
   \label{fig:visual_div2k}
\end{figure*}

\subsection{Uncertainty-Driven Diffusion Guidance}
\label{sec:guidance}

In this section, we describe how we use the estimated uncertainty to guide the sampling process. We start from the guidance proposed by DiffBIR~\cite{dadiff}, which is given by
\begin{align}
    \mathcal{L}_\text{diffbir} &= \left\| (\mathbf{1}-\tanh (\mathcal{G}(\mathbf{I}_\text{rm}))) \odot (\mathcal{D}(\hat{\mathbf{z}}_{0|t}) - \mathbf{I}_\text{rm}) \right\|^2_2, \label{eq:diffbir} \\
    \mathbf{z}_{t-1} &\sim \mathcal N(\mu_\theta(\mathbf{z}_{t}, \hat{\mathbf{z}}_{0|t})-s\nabla_{\hat{\mathbf{z}}_{0|t}}\mathcal{L}_\text{diffbir}, \sigma_t^2 \mathbf{I}) \label{eq:guidance_base},
\end{align}
where $\mathcal{G}$ is an image gradient operator, $\mu_\theta$ is conditioned on $\bm{\mu} = \mathcal{E}(\mathbf{I}_\text{rm})$, and $s$ is the guidance scale. This guidance lets the sampling process move towards $\mathbf{I}_\text{rm}$ for low-textured areas and thus enhances the fidelity.

Our core idea is to make this guidance more adaptive according to the estimated uncertainty. Specifically, 
we form a product of Gaussians by multiplying the estimated uncertainty distribution by the posterior in Eq.~(\ref{eq:reverse}), then obtain its maximizer by
\begin{align}
    \mathbf{z}^\star_{t-1} &= \argmax\limits_{\mathbf{z}_{t-1}} p(\mathbf{z}_{t-1}|\bm{\mu}, \mathbf{\Sigma}'_\text{u}) \cdot p(\mathbf{z}_{t-1}|\mathbf{z}_t, \hat{\mathbf{z}}_{0|t}) \nonumber \\
    &= \argmax\limits_{\mathbf{z}_{t-1}} \mathcal{N}(\mathbf{z}_{t-1}; \sqrt{\bar{\alpha}_{t-1}}\bm{\mu}, \bar{\alpha}_{t-1} \diag (\mathbf{\Sigma}'_\text{u})) \nonumber \\
    &\quad\quad\cdot \mathcal{N}(\mathbf{z}_{t-1}; \mu_\theta(\mathbf{z}_t, \hat{\mathbf{z}}_{0|t}), \sigma^2_t \mathbf{I}) \nonumber \\
    &= \frac{\sigma^2_t \cdot \sqrt{\bar{\alpha}_{t-1}}\bm{\mu} + \bar{\alpha}_{t-1}\mathbf{\Sigma}'_\text{u}\cdot \mu_\theta(\mathbf{z}_t, \hat{\mathbf{z}}_{0|t})}{\sigma^2_t \mathbf{1} + \bar{\alpha}_{t-1}\mathbf{\Sigma}'_\text{u}},
\label{eq:z_star}
\end{align}
where $\mathbf{\Sigma}'_\text{u}=\gamma\mathbf{\Sigma}_\text{u}$ and $\gamma \in \mathbb{R}^+$ is the uncertainty scale to control the perception-distortion balance (lower for better fidelity), and $\mathbb{E}[\mathbf{z}_{t-1}|\bm{\mu}]=\sqrt{\bar{\alpha}_{t-1}}\bm{\mu}$ is based on Eq.~(\ref{eq:forward}). Finally, our diffusion guidance is given by
\begin{align}
    \mathcal{L}_\text{ugdiff} &= \left\| \sqrt{\bar{\alpha}_{t-1}}\hat{\mathbf{z}}_{0|t} - \mathbf{z}_{t-1}^\star \right\|^2_2,
\label{eq:guidance}
\end{align}
which is substituted into Eq.~(\ref{eq:guidance_base}) to replace $\mathcal{L}_\text{diffbir}$, making the sampling process move towards $\mathbf{z}_{t-1}^\star$ instead of $\mathbf{I}_\text{rm}$.

\begin{figure}[t]
  \centering
  \includegraphics[width=0.75\linewidth]{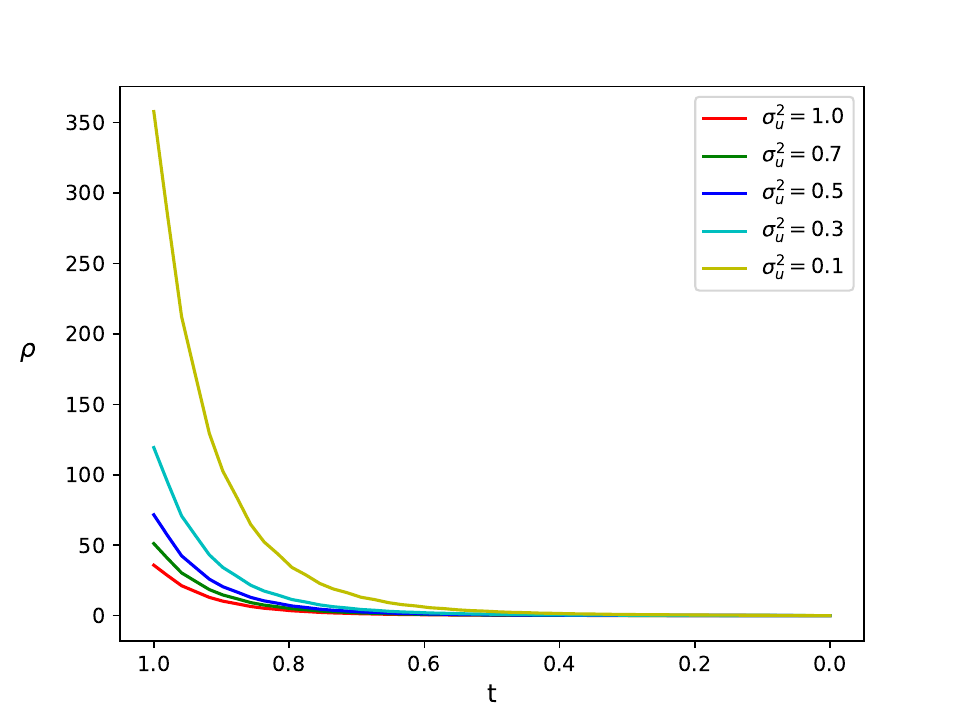}
  \caption{Relative ratio of weighting over time under different uncertainties $\sigma^2_\text{u}$. The ratio is defined by $\rho=\sigma^2_t / (\bar{\alpha}_{t-1}\sigma^2_\text{u})$. Note that the curves are depicted using the scaled linear noise scheduler with $\beta_1=8.5\times10^{-4}$ and $\beta_T=1.2\times10^{-2}$.}
  \label{fig:weight_ratio}
\end{figure}

It is worth noting that Eq.~(\ref{eq:z_star}) actually means that we perform a linear combination on the mean of the two distributions. As we can see in Fig.~\ref{fig:weight_ratio}, the relative ratio of the weighting $\rho=\sigma^2_t / (\bar{\alpha}_{t-1} \sigma^2_\text{u})$ increases as the uncertainty $\sigma^2_\text{u}$ decreases, causing the sampling to go towards the direction of $\bm{\mu}$. Furthermore, $\rho$ is typically very large at the early steps but decreases faster later, allowing our method to relax reliance on $\bm{\mu}$ at the later steps and improve perceptual quality.

\begin{table}[t]
    \caption{Quantitative comparisons of $4 \times$ upsampling. \underline{Underline} indicates superior performance between DiffBIR and ours at comparable PSNR levels. \textcolor{red}{Red} and \textcolor{blue}{blue} respectively indicate the best and the second best performance among all methods. We set $s=100$ for all of our methods.}
    \centering
    \vspace{0.7em}
    \resizebox{1.0\columnwidth}{!}{
    \begin{tabular}{llcccc}
        \toprule
        Dataset & Method & PSNR $\uparrow$ & SSIM $\uparrow$ & LPIPS $\downarrow$ & NIQE $\downarrow$ \\
        \midrule
        \multirow{12}{*}{DIV2K-Val} & StableSR~\cite{stablesr} & 21.6100 & 0.5267 & 0.3113 & 4.7581 \\
        & ResShift~\cite{resshift} & 22.6640 & \textcolor{red}{0.5884} & \textcolor{blue}{0.3077} & 6.9142 \\
        & SeeSR~\cite{seesr} & 21.9703 & 0.5592 & 0.3194 & 4.8097 \\
        & FaithDiff~\cite{faithdiff} & 21.7965 & 0.5348 & 0.3118 & 4.9149 \\
        & SUPIR~\cite{supir} & 21.4792 & 0.5005 & 0.3625 & 6.3373 \\
        & PiSA-SR~\cite{pisa-sr} & 22.2189 & 0.5624 & \textcolor{red}{0.2823} & \textcolor{blue}{4.5577} \\
        \cmidrule(lr){2-6}
        & DiffBIR~\cite{diffbir} ($s=0.0$) & 21.4790 & 0.4969 & 0.3670 & 4.9934 \\
        & DiffBIR~\cite{diffbir} ($s=0.15$) & 22.2730 & 0.5418 & 0.3413 & 5.2346 \\
        & DiffBIR~\cite{diffbir} ($s=0.5$) & \textcolor{red}{22.7957} & 0.5707 & \underline{0.3419} & 5.6970 \\
        & Ours ($\gamma=0.5$) & 21.5034 & \underline{0.5087} & \underline{0.3510} & \textcolor{red}{\underline{4.1650}} \\
        & Ours ($\gamma=0.1$) & 22.2350 & \underline{0.5471} & \underline{0.3406} & \underline{4.7852} \\
        & Ours ($\gamma=0.01$) & \textcolor{blue}{22.7391} & \textcolor{blue}{\underline{0.5737}} & 0.3430 & \underline{5.3038} \\
        \midrule
        \multirow{12}{*}{RealSR} & StableSR~\cite{stablesr} & 23.1258 & 0.6722 & 0.3002 & 5.8812 \\
        & ResShift~\cite{resshift} & 23.9576 & 0.7049 & 0.3279 & 8.0707 \\
        & SeeSR~\cite{seesr} & 23.5972 & 0.6872 & 0.3007 & 5.3957 \\
        & FaithDiff~\cite{faithdiff} & 23.7191 & 0.6721 & \textcolor{blue}{0.2887} & \textcolor{blue}{5.3895} \\
        & SUPIR~\cite{supir} & 23.4637 & 0.6326 & 0.3716 & 7.5991 \\
        & PiSA-SR~\cite{pisa-sr} & 23.9644 & \textcolor{red}{0.7089} & \textcolor{red}{0.2672} & 5.5014 \\
        \cmidrule(lr){2-6}
        & DiffBIR~\cite{diffbir} ($s=0.0$) & 23.3298 & 0.6134 & 0.3652 & 5.8397 \\
        & DiffBIR~\cite{diffbir} ($s=0.3$) & 24.0969 & \underline{0.6707} & 0.3279 & 6.2908 \\
        & DiffBIR~\cite{diffbir} ($s=1.0$) & \textcolor{red}{24.6003} & \textcolor{blue}{\underline{0.7058}} & 0.3105 & 6.8355 \\
        & Ours ($\gamma=0.5$) & 23.4219 & \underline{0.6301} & \underline{0.3324} & \textcolor{red}{\underline{4.9070}} \\
        & Ours ($\gamma=0.1$) & 24.1528 & 0.6700 & \underline{0.3170} & \underline{5.6148} \\
        & Ours ($\gamma=0.01$) & \textcolor{blue}{24.5990} & 0.6946 & \underline{0.3072} & \underline{6.0928} \\
        \bottomrule
    \end{tabular}
    }
    \label{tab:quantitative}
\end{table}

\section{Experiments}

\subsection{Experiment Settings}
\label{sec:setting}

\heading[0ex]{Data preparation.} We use the test datasets from StableSR~\cite{stablesr}: a synthetic dataset \emph{DIV2K-Val}~\cite{div2k} and a real dataset \emph{RealSR}~\cite{realsr}, both of which involve $4\times$ upsampling from $128 \times 128$ to $512 \times 512$. We train our uncertainty estimation model $\mathcal{E}'$ on LSDIR~\cite{lsdir} and the first 10K face images from FFHQ~\cite{ffhq}, where the LQ-HQ pairs are generated by the degradation pipeline from Real-ESRGAN~\cite{realesr_gan}. We also randomly generate 100 crops from DIV2K~\cite{div2k} for validation, denoted by \emph{DIV2K-Ours}.

\heading{Implementation details.} We build our method upon the DiffBIR~\cite{dadiff} baseline. We use the ControlNet~\cite{controlnet} model from DiffBIR, and the pretrained diffusion and VAE~\cite{vae} models from Stable Diffusion v2.1~\cite{sd}. For the text encoder, we use the same negative prompts as DiffBIR and keep the positive prompts empty. We also adopt BSRNet~\cite{bsrgan} as our restoration model $\mathcal{R}$. We use the VAE architecture and train from scratch our uncertainty estimation model $\mathcal{E}'$ using AdamW~\cite{adamw} with a learning rate of $10^{-4}$, a batch size of 128, and a crop size of $256\times256$ on 4 NVIDIA A100 GPUs for 90K iterations. For sampling, we use the spaced DDPM~\cite{iddpm} sampler with the scaled linear noise scheduler, where $\beta_1=8.5\times10^{-4}$ and $\beta_T=1.2\times10^{-2}$, for totally $T=50$ steps.

\heading{Methods in comparison.} We compare our methods with the state-of-the-art diffusion-based SR methods, including StableSR~\cite{stablesr}, ResShift~\cite{resshift}, SeeSR~\cite{seesr}, FaithDiff~\cite{faithdiff}, SUPIR~\cite{supir}, PiSA-SR~\cite{pisa-sr}, and DiffBIR~\cite{diffbir}. We provide all the numbers using the official code with default settings. 

\heading{Evaluation metrics.} There are 4 metrics used for evaluation. PSNR and SSIM are reference-based metrics for fidelity. LPIPS~\cite{lpips} is a reference-based metric, and NIQE~\cite{niqe} is a non-reference metric, both for perceptual quality. 

\subsection{Ablation Studies}
\label{sec:analysis}

\heading[0ex]{Ablation study on mean estimation.}
To validate the importance of Eq.~(\ref{eq:loss_uncertainty}), we train an uncertainty estimation model $\mathcal{E}_\text{nll}$ using Eq.~(\ref{eq:loss_nll}) and employ its mean estimation. Table~\ref{tab:uncertainty} shows that using the mean estimation from BSRNet yields superior performance across all metrics on DIV2K-Ours, thereby justifying our focus on accurate uncertainty estimation.

\heading{Ablation study on guidance parameters.} Table~\ref{tab:guidance_parameter} shows the ablation study of guidance parameters on DIV2K-Ours. In DiffBIR, fidelity improves while perceptual quality decreases as the guidance scale $s$ increases, but our method exhibits the opposite trend. This demonstrates a fundamental distinction between our method and DiffBIR: rather than merely pulling the results towards the restored image $\mathbf{I}_\text{rm}$, we alter the convergence characteristics of the sampling process. Therefore, we fix $s=100$ because it achieves the optimal balance among all configurations with $\gamma=1.0$. With $s$ fixed at $100$, different values of $\gamma$ offer different trade-offs. Specifically, a smaller $\gamma$ results in a stronger pull towards the mean $\bm{\mu}$, thereby improving fidelity, and vice versa. Given its clear influence on the perception-distortion trade-off, we adopt $\gamma$ as the primary parameter to control this balance throughout our experiments.

\subsection{Comparisons with State-of-the-Arts}
\label{sec:sota}

\heading[0ex]{Quantitative comparisons.} Table~\ref{tab:quantitative} shows the quantitative comparisons. Given that DiffBIR serves as our baseline, we first compare our method against it at comparable PSNR levels. As demonstrated in the table, our method consistently achieves superior perceptual quality. Furthermore, our method performs favorably against other state-of-the-art methods, particularly in terms of NIQE scores.

Fig.~\ref{fig:tradeoff} shows the perception-distortion trade-off of diffusion-based SR methods on RealSR using PSNR and NIQE. As we can see in the figure, our method achieves the best perception-distortion balance among all of the compared methods.

\heading{Qualitative comparisons.} Fig.~\ref{fig:visual_div2k} shows the qualitative comparisons on DIV2K-Val and RealSR. Our method ($\gamma=0.5$) yields visually similar results to DiffBIR ($s=0$) when their PSNR values are comparable. If we decrease $\gamma$ from $0.5$ to $0.1$, we can avoid unnecessary high-frequency details particularly in flat regions, showing the effectiveness of our guidance method. Regarding other methods, PiSA-SR, SeeSR, and FaithDiff often introduce details in undesired regions, while ResShift's results lack high-frequency details.

\begin{table}
    \caption{Ablation study on mean estimation. \textbf{Bold} indicates superior performance. We conduct the experiments using DiffBIR~\cite{diffbir} without guidance ($s=0$).}
    \centering
    \vspace{0.7em}
    \resizebox{0.75\columnwidth}{!}{
    \begin{tabular}{ccccc}
        \toprule
        $\mathbf{I}_\text{rm}$ & PSNR $\uparrow$ & SSIM $\uparrow$ & LPIPS $\downarrow$ & NIQE $\downarrow$ \\
        \midrule
        $\text{BSRNet}(\mathbf{I}_\text{lq})$ & \textbf{20.2307} & \textbf{0.4222} & \textbf{0.4388} & \textbf{4.6974} \\
        $\mathcal{D}(\mathcal{E}_\text{nll}(\mathbf{I}_\text{lq}))$ & 19.5111 & 0.3844 & 0.4854 & 5.6106 \\
        \bottomrule
    \end{tabular}
    }
    \label{tab:uncertainty}
\end{table}

\begin{table}
    \caption{Ablation study on guidance parameters. \textbf{Bold} indicates the best performance.}
    \centering
    \vspace{0.7em}
    \resizebox{0.75\columnwidth}{!}{
    \begin{tabular}{cccccc}
        \toprule
        $s$ & $\gamma$ & PSNR $\uparrow$ & SSIM $\uparrow$ & LPIPS $\downarrow$ & NIQE $\downarrow$ \\
        \midrule
        1 & 1.0 & \textbf{20.2278} & \textbf{0.4222} & 0.4387 & 4.6801 \\
        10 & 1.0 & 20.2040 & 0.4219 & 0.4371 & 4.5573 \\
        100 & 1.0 & 19.8139 & 0.4113 & \textbf{0.4271} & \textbf{3.7287} \\
        200 & 1.0 & 18.9496 & 0.3779 & 0.4628 & 4.0655 \\
        \midrule
        100 & 1.0 & 19.8139 & 0.4113 & 0.4271 & \textbf{3.7287} \\
        100 & 0.5 & 20.2100 & 0.4340 & 0.4185 & 3.8541 \\
        100 & 0.1 & 20.9637 & 0.4780 & \textbf{0.4095} & 4.5425 \\
        100 & 0.01 & \textbf{21.4769} & \textbf{0.5085} & 0.4134 & 5.2745 \\
        \bottomrule
    \end{tabular}
    }
    \label{tab:guidance_parameter}
\end{table}

\section{Conclusion}

In this paper, we have presented a method for training an uncertainty estimation model in the latent space and leveraging this uncertainty to guide the diffusion process. The experimental results have demonstrated that our proposed method strikes a better perception-distortion balance than state-of-the-art methods. For future work, distilling the diffusion process towards single-step or extending our method to a broader range of tasks could be possible directions.

\bibliographystyle{IEEEtran}
\bibliography{main}

\appendix
\onecolumn
\begin{center}
    {\large \bf \MakeUppercase{Uncertainty-Guided Latent Diffusion Models \\ for Faithful Super Resolution} \par}
    \vspace{0.5em}
    {\large Supplementary Material \par}
    \vspace{1.0em}
\end{center}
\section{Ablation Study on Parameterization}

Table~\ref{tab:parameterization} shows the ablation study of the parameterization of our uncertainty estimation model on DIV2K-Ours. By parameterization, we mean whether to use $\sigma_\text{u}$ or $\sigma^2_\text{u}$ as the output of the model. The parameterization with $\sigma_\text{u}$ performs better in terms of most metrics, so we choose it as our final solution, \ie, $\mathbf{\Sigma}_\text{u}=\mathcal{E}'(\mathbf{I}_\text{rm}) \odot \mathcal{E}'(\mathbf{I}_\text{rm})$.

\begin{table}[h]
    \caption{Ablation study on parameterization. \textbf{Bold} indicates the better performance. We set $s=100$ and $\gamma=0.1$ for the experiments.}
    \centering
    \vspace{1em}
    \resizebox{0.4\columnwidth}{!}{
    \begin{tabular}{ccccc}
        \toprule
        Param. & PSNR $\uparrow$ & SSIM $\uparrow$ & LPIPS $\downarrow$ & NIQE $\downarrow$ \\
        \midrule
        $\sigma^2_\text{u}$ & 20.7987 & 0.4676 & 0.4118 & \textbf{4.3496} \\
        $\sigma_\text{u}$ & \textbf{20.9637} & \textbf{0.4780} & \textbf{0.4095} & 4.5425 \\
        \bottomrule
    \end{tabular}
    }
    \label{tab:parameterization}
\end{table}

\section{Visual Comparison on Uncertainty Scale}

\begin{figure*}[h]
  \centering
   \includegraphics[width=0.75\linewidth]{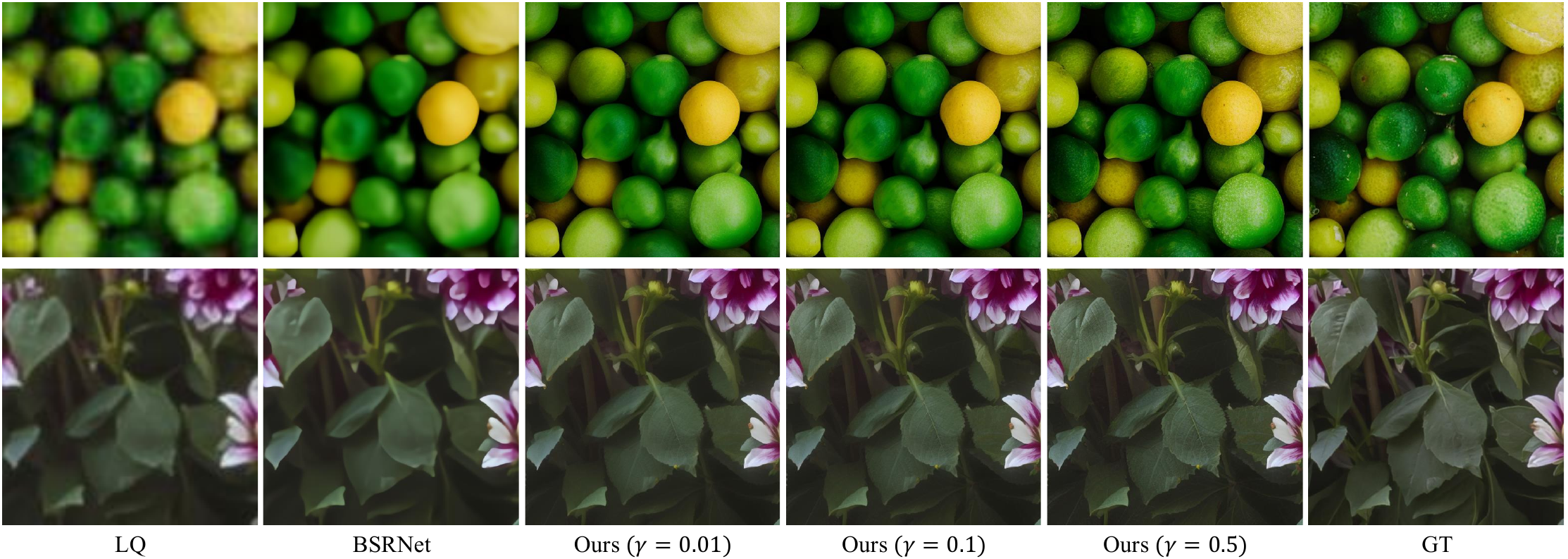}
   \caption{Visual comparison under different uncertainty scales on DIV2K-Val (upper) and RealSR (bottom). We set $s=100$ for all of our methods.}
   \label{fig:gamma}
\end{figure*}

We provide a comparison on the uncertainty scale $\gamma$ in Fig.~\ref{fig:gamma}. As shown in the upper part of the figure, a higher value of $\gamma$ is preferred when richer high-frequency details are desired. In contrast, we can also use a lower $\gamma$ to avoid unnecessary details, making the result look more like the restoration model BSRNet~\cite{bsrgan}, as in the bottom part.

\section{Impact of Guidance on the Sampling Process}

\begin{figure*}[h]
  \centering
   \includegraphics[width=0.9\linewidth]{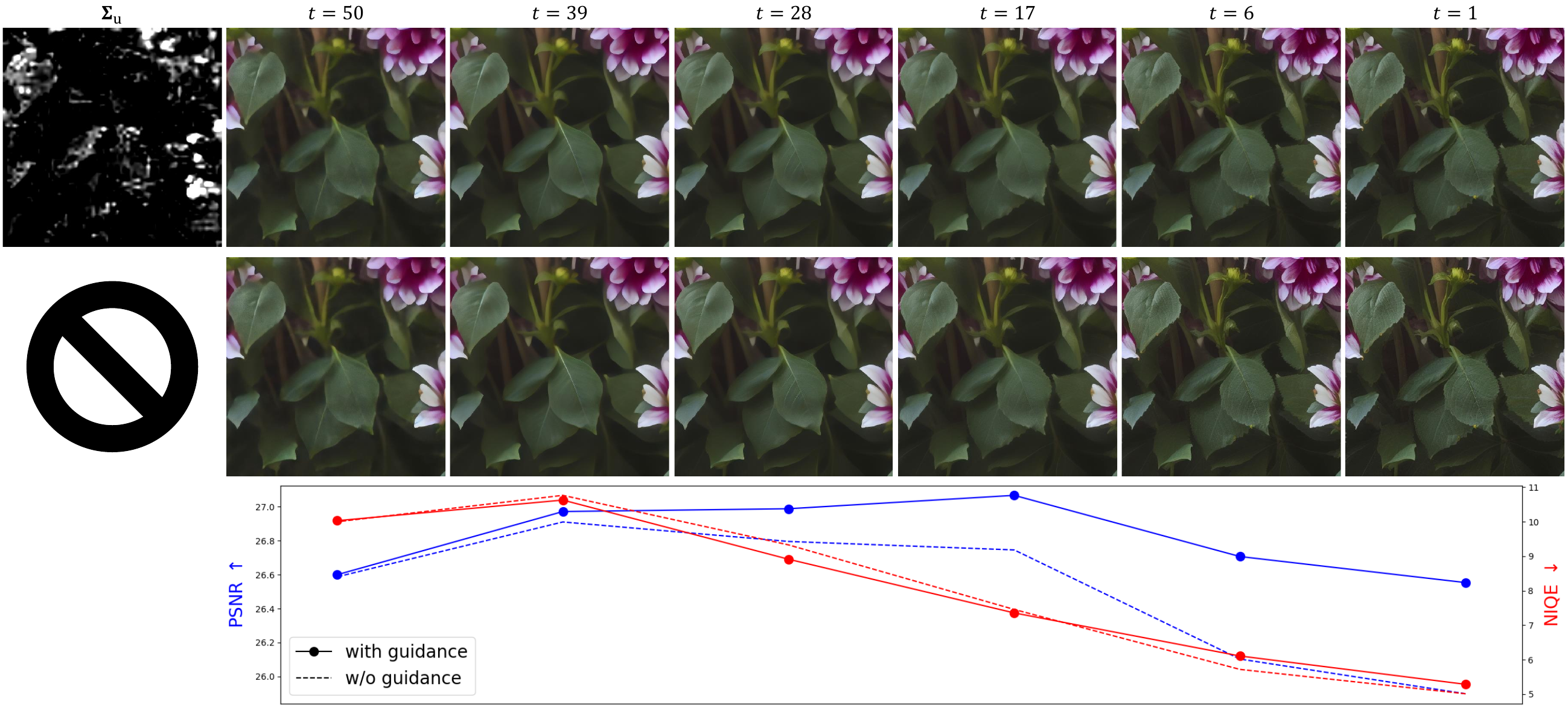}
   \caption{Visualization and analysis of the sampling process with and w/o our guidance. We set $s=100$ and $\gamma=0.1$ for the guidance.}
   \label{fig:sampling_process}
\end{figure*}

Fig.~\ref{fig:sampling_process} presents a qualitative and quantitative comparison of the sampling process, both with and without the proposed guidance. Recall that there are a total of 50 sampling steps; we select 6 steps within this range to visualize the corresponding $\mathcal{D}(\hat{\mathbf{z}}_{0|t})$. As illustrated in the figure, our proposed guidance suppresses unnecessary textures in low-uncertainty regions, thereby achieving higher PSNR eventually. Furthermore, the NIQE curve with our guidance is nearly identical to the baseline, demonstrating that our adaptive mechanism, which relaxes the reliance on $\bm{\mu}$ at later sampling steps, effectively preserves perceptual quality.

\section{More Qualitative Comparisons with State-of-the-Arts}

\begin{figure*}[t]
  \centering
   \includegraphics[width=\linewidth]{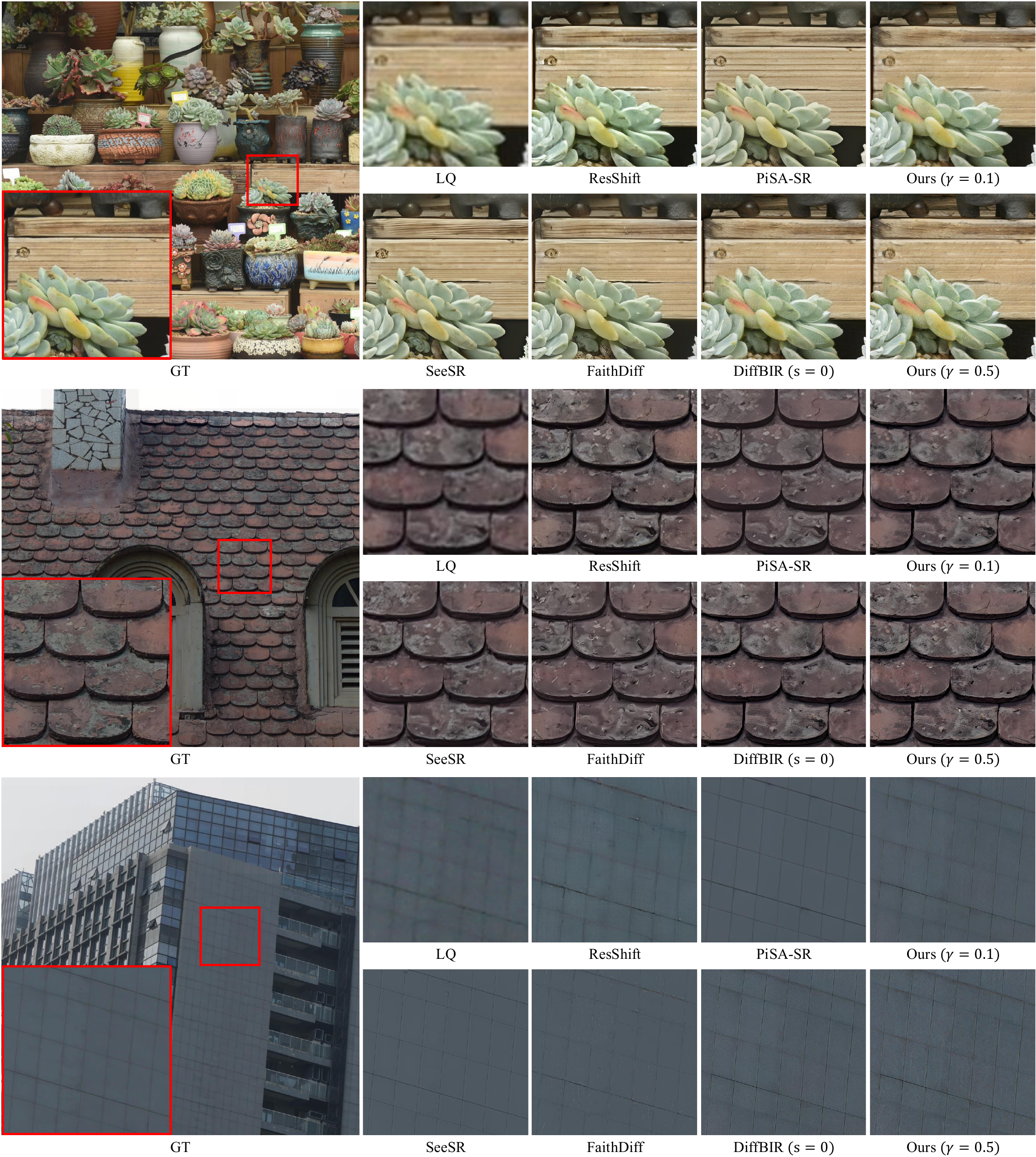}
   \caption{Qualitative comparison of $4\times$ upsampling on DRealSR. We set $s=100$ for all of our methods.}
   \label{fig:visual_drealsr_suppl}
\end{figure*}

\begin{figure*}[t]
  \centering
   \includegraphics[width=\linewidth]{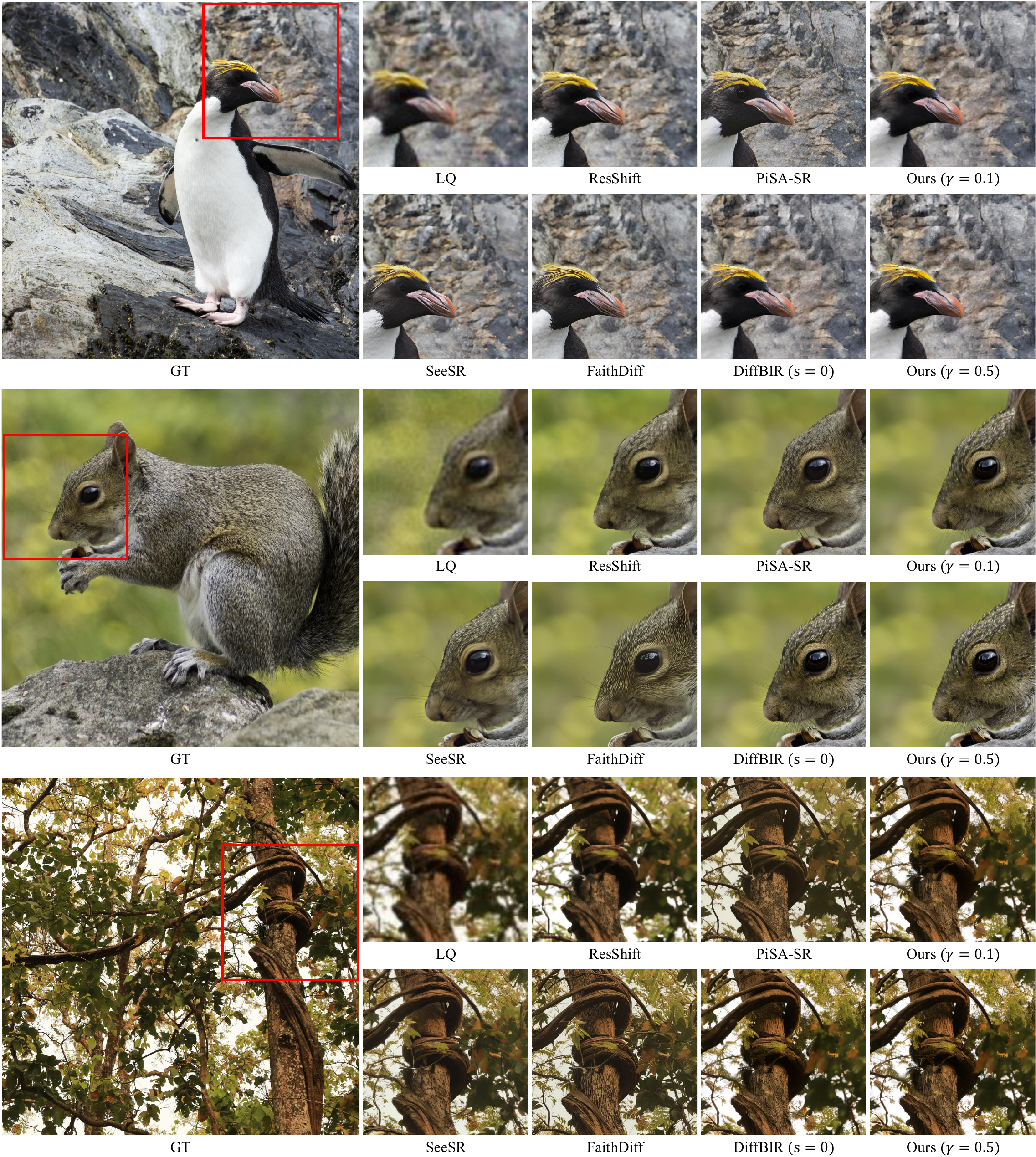}
   \caption{Qualitative comparison of $4\times$ upsampling on DIV2K-Val. We set $s=100$ for all of our methods.}
   \label{fig:visual_div2k_suppl}
\end{figure*}

We provide more qualitative comparisons with state-of-the-arts in Figs.~\ref{fig:visual_drealsr_suppl}--\ref{fig:visual_div2k_suppl}, where Fig.~\ref{fig:visual_drealsr_suppl} uses the \emph{DRealSR}~\cite{drealsr} dataset from StableSR~\cite{stablesr}. Consistent with the observations in Fig. 3 of the main paper, our method performs similiarly to DiffBIR~\cite{diffbir} with $\gamma=0.5$. By setting $\gamma=0.1$, we can remove unnecessary texture details, and hence achieve a better perception-distortion balance than the compared methods.

\section{Proofs}

\subsection{Equation~(\ref{eq:var_star}): Optimal Uncertainty}

If $\mathbf{z}_\text{gt}$ and $\bm{\mu}$ are given, the point-wise term in $\mathcal{L}_\text{nll}$ is deterministic, and we can set the gradient to zero as
\begin{align}
    \frac{\partial \mathcal{L}_\text{nll}}{\partial \mathbf{\Sigma}_{\text{u}}} &= \frac{\partial}{\partial \mathbf{\Sigma}_{\text{u}}} \left[ \sum\limits_{i=1}^d \frac{( \mathbf{z}_{\text{gt}, i} - {\bm{\mu}}_i )^2}{2\mathbf{\Sigma}_{\text{u},i}} + \frac{1}{2}\ln\mathbf{\Sigma}_{\text{u},i} \right] \nonumber \\
    &= \left[ -\frac{( \mathbf{z}_{\text{gt}, i} - {\bm{\mu}}_i )^2}{2\mathbf{\Sigma}^2_{\text{u}, i}} + \frac{1}{2{\mathbf{\Sigma}}_{\text{u},i}}\right]_{i=1}^d \nonumber \\
    &= \mathbf{0} \nonumber.
\end{align}
Therefore, we have 
\begin{align}
    \mathbf{\Sigma}^\star_\text{u} &= \argmin\limits_{\mathbf{\Sigma}_\text{u}} \mathcal{L}_\text{nll} = (\mathbf{z}_\text{gt} - \bm{\mu}) \odot (\mathbf{z}_\text{gt} - \bm{\mu}) \nonumber.
\end{align}

\subsection{Equation~(\ref{eq:z_star}): Maximizer of the Product of Gaussians}

To calculate
\begin{align}
    \mathbf{z}^\star_{t-1} &= \argmax\limits_{\mathbf{z}_{t-1}} \mathcal{N}(\mathbf{z}_{t-1}; \sqrt{\bar{\alpha}_{t-1}}\bm{\mu}, \bar{\alpha}_{t-1} \diag(\mathbf{\Sigma}'_\text{u})) \cdot \mathcal{N}(\mathbf{z}_{t-1}; \mu_\theta(\mathbf{z}_t, \hat{\mathbf{z}}_{0|t}), \sigma^2_t \mathbf{I}) \nonumber \\
    &= \argmin\limits_{\mathbf{z}_{t-1}} \sum\limits_{i=1}^d \frac{( \mathbf{z}_{t-1, i}-\sqrt{\bar{\alpha}_{t-1}}{\bm{\mu}}_i )^2}{2 \bar{\alpha}_{t-1} \mathbf{\Sigma}'_{\text{u}, i}} + \frac{( \mathbf{z}_{t-1, i}-\mu_{\theta, i}(\mathbf{z}_t, \hat{\mathbf{z}}_{0|t}) )^2}{2 \sigma^2_t} \nonumber,
\end{align}
we can set the gradient to zero as
\begin{align}
    &\quad \frac{\partial}{\partial \mathbf{z}_{t-1}} \left[ \sum\limits_{i=1}^d \frac{( \mathbf{z}_{t-1, i}-\sqrt{\bar{\alpha}_{t-1}}{\bm{\mu}}_i )^2}{2 \bar{\alpha}_{t-1} \mathbf{\Sigma}'_{\text{u}, i}} + \frac{( \mathbf{z}_{t-1, i}-\mu_{\theta, i}(\mathbf{z}_t, \hat{\mathbf{z}}_{0|t}) )^2}{2 \sigma^2_t} \right] \nonumber \\
    &= \mathbf{z}_{t-1} \left( \frac{1}{\bar{\alpha}_{t-1} \mathbf{\Sigma}'_\text{u}} + \frac{1}{\sigma^2_t \mathbf{1}} \right) - \left( \frac{\sqrt{\bar{\alpha}_{t-1}}\bm{\mu}}{\bar{\alpha}_{t-1} \mathbf{\Sigma}'_\text{u}} + \frac{\mu_\theta(\mathbf{z}_t, \hat{\mathbf{z}}_{0|t})}{\sigma^2_t \mathbf{1}} \right) \nonumber \\
    &= \mathbf{0} \nonumber.
\end{align}
Therefore, we have
\begin{align}
    \mathbf{z}^\star_{t-1} = \frac{\sigma^2_t \cdot \sqrt{\bar{\alpha}_{t-1}}\bm{\mu} + \bar{\alpha}_{t-1}\mathbf{\Sigma}'_\text{u}\cdot \mu_\theta(\mathbf{z}_t, \hat{\mathbf{z}}_{0|t})}{\sigma^2_t \mathbf{1} + \bar{\alpha}_{t-1}\mathbf{\Sigma}'_\text{u}} \nonumber.
\end{align}

\end{document}